\documentclass[journal,twoside,web]{ieeecolor2}

\usepackage{generic}
\usepackage{cite}
\usepackage{amsmath,amssymb,amsfonts}
\usepackage{algorithmic}
\usepackage{textcomp}
\usepackage{graphicx} 
\usepackage{graphicx,amssymb,amstext,amsmath}
\usepackage{multirow}
\usepackage{subcaption}
\usepackage[ruled,vlined]{algorithm2e}
\usepackage{bbm}
\usepackage{epstopdf}
\usepackage{amsfonts} 
\usepackage{color}
\usepackage{amssymb}
\usepackage{bm,nicefrac}
\usepackage{amstext}
\usepackage{ntheorem}

\usepackage{mathtools}
\usepackage{algorithmic}
\usepackage[font=small]{caption} 
\usepackage{orcidlink}
\usepackage{lineno,hyperref}
\usepackage{booktabs}
\usepackage{adjustbox} 
\usepackage{tablefootnote}
\hypersetup{
colorlinks,
linkcolor={red!50!black},
citecolor={blue!50!black},
urlcolor={blue!80!black}
}
\def\BibTeX{{\rm B\kern-.05em{\sc i\kern-.025em b}\kern-.08em
    T\kern-.1667em\lower.7ex\hbox{E}\kern-.125emX}}
\begin{document}
\title{Conservative Subject Invariant EMG-based Gesture Recognition}
\author{Hamed Rafiei${\orcidlink{0000-0002-0755-9814}}$ and Ali Mousavi${\orcidlink{0000-0002-4084-6102}}$
\thanks{H. Rafiei is with Department of Electrical Engineering, Center of Excellence on Soft Computing and Intelligent Information Processing, Ferdowsi University of Mashhad, Mashhad, Iran (e-mail: rafiei.hamed@mail.um.ac.ir).}
\thanks{A. Mousavi is with Department of Computer Engineering, Ne. C., Islamic Azad University, Neyshabur, Iran (e-mail: mousavi@iau.ac.ir).}}

\maketitle

\begin{abstract}
Cross-subject generalization remains a fundamental challenge in surface electromyography (sEMG)-based gesture recognition. Although deep learning methods have improved within-subject performance, they often rely on subject-specific data and struggle to balance invariance and discriminability. In this work, we propose a conservative multi-objective learning framework for subject-invariant sEMG gesture recognition. The proposed model adopts a multi-head architecture that jointly optimizes gesture classification, adversarial subject confusion through gradient reversal, and triplet-based metric learning to encourage discriminative and subject-invariant representations. To improve optimization stability, a Lipschitz-inspired adaptive weighting mechanism is introduced to dynamically balance the auxiliary objectives according to their relative magnitudes during training. The proposed method is evaluated on two benchmark datasets: UCI EMG (36 subjects, 6 gestures) and NinaPro DB5 (10 subjects, 10 gestures). On the UCI EMG dataset, the method achieves 84.48\% accuracy compared to 78.2\% reported by state-of-the-art methods. On NinaPro DB5, it achieves 61.44\% accuracy versus 41.30\%, corresponding to a 49\% relative improvement. In addition, the proposed framework reduces cross-subject prediction variance and produces more structured latent representations. These results indicate that jointly enforcing invariance and discriminability through adaptive multi-objective optimization leads to more stable training and improved cross-subject generalization in sEMG-based gesture recognition systems.
\end{abstract}


\section{Introduction}
\IEEEPARstart{S}{urface} electromyography (sEMG) captures the electrical activity of muscle fibers and motor neurons through non-invasive sensors, providing rich insight into muscle activation during movement~\cite{oskoei2007myoelectric}. In particular, sEMG-based hand gesture recognition seeks to decode these signals to identify discrete hand movements, enabling intuitive control of robotic systems, prosthetic devices, and assistive technologies—especially for individuals with motor impairments. Owing to its non-invasive nature and high temporal resolution, sEMG has been widely adopted in applications such as human--computer interaction, virtual and augmented reality, rehabilitation, and wearable robotics \cite{ge2023gesture,jaramillo2020real,zhang2024online}. Despite these advances, achieving reliable performance across different users and recording conditions remains a fundamental challenge. In practice, models must learn representations that generalize beyond subject-specific characteristics and session-dependent variations, which often limits their robustness in real-world deployments.

sEMG signals are inherently complex and highly variable \cite{rafiei2025understandable}, exhibiting strong sensitivity to both inter-subject differences \cite{liu2015towards} and inter-session variability \cite{du2017surface}. These variations arise from intrinsic factors, such as differences in muscle anatomy and physiology across individuals, as well as extrinsic factors, including electrode displacement, variations in skin--electrode contact, and changes in muscle condition across recording sessions. As a result, sEMG feature representations can vary significantly even when different subjects perform the same gesture under nominally identical conditions. This pronounced subject-dependent variability poses a major obstacle to cross-subject generalization, as models trained on a limited set of users often capture subject-specific patterns rather than task-relevant muscle activation dynamics. Notably, the individuality encoded in sEMG signals is so strong that it has been successfully leveraged for biometric identification, further underscoring the challenge of separating gesture-related information from subject-specific signatures \cite{pradhan2021performance,jiang2022optimization,yang2024emgbench,fan2024surface}.


Several strategies have been explored to address inter-subject variability in sEMG-based gesture recognition, with increasing emphasis on learning subject-invariant representations. Early approaches relied on carefully designed handcrafted features to improve robustness across users and recording conditions~\cite{jiang2022optimization,phinyomark2013emg}. However, the inherently strong subject-specific nature of sEMG signals limits their effectiveness. Subsequently, transfer learning and domain adaptation methods were introduced to mitigate cross-subject performance degradation by adapting models trained on source subjects to a target user~\cite{fan2023improving,wang2023iterative}. While these approaches can improve performance, they typically require access to target-subject data and may suffer from error accumulation when pseudo-labels become unreliable under large inter-subject distribution shifts. More recent work has explored adversarial learning as a means of encouraging subject-invariant representations without explicit target-domain supervision. By introducing a subject discriminator and training the encoder to confuse it, these methods aim to suppress subject-identifying information in the latent space. Representative examples include multi-source adversarial alignment and feature disentanglement approaches that explicitly promote invariance across subjects~\cite{su2025multi,fan2024surface}. However, adversarial objectives alone do not explicitly enforce separation between gesture classes and may lead to over-alignment or class overlap in the latent space. Recent benchmark studies, such as EMGBench~\cite{yang2024emgbench}, evaluate a broad range of models and training strategies under cross-subject and cross-session settings, and show that performance remains limited under strong distribution shifts. These findings highlight the difficulty of learning representations that generalize across subjects and suggest that invariance alone is insufficient without strong task-discriminative constraints. Recent work has explored Lipschitz-based constraints to improve robustness of sEMG classifiers to input perturbations~\cite{neacsu2024lipschitz}; however, such approaches primarily focus on stability and do not explicitly address cross-subject variability. Moreover, Lipschitz continuity has been studied as a principled framework for controlling gradient behavior and improving training stability in deep networks~\cite{tsuzuku2018lipschitz,gouk2021regularisation}. In this context, it provides a useful interpretation of how the scale of a loss function relates to the magnitude of its gradients during optimization. These insights motivate the use of adaptive mechanisms to balance the influence of adversarial and discriminative objectives in multi-objective training.


Alongside domain adaptation and adversarial strategies, deep learning models have been widely adopted to exploit the rich temporal and spectral characteristics of sEMG signals. Convolutional neural networks (CNNs) have been used to automatically extract discriminative features from both time and frequency domains~\cite{gao2022multifeatured,shen2023ica}. In addition, hybrid architectures that combine CNNs with recurrent neural networks (RNNs) have been proposed to more effectively capture the spatiotemporal dependencies in muscle activation patterns~\cite{zhang2020learning,chen2020hand,wang2025residual}. Although these models significantly enhance representation capacity and recognition accuracy, increased expressive power alone does not guarantee robustness to subject variability. In practice, deep models often learn latent representations in which gesture-related information is entangled with subject-specific characteristics, limiting their ability to generalize across users~\cite{cote2020interpreting}. To address this limitation, recent work has explored explicit disentanglement strategies aimed at separating task-relevant and subject-dependent factors in the latent space. For instance, Fan \textit{et al.}~\cite{fan2024surface} proposed a multi-encoder architecture with a shared decoder to decompose sEMG features into gesture-related and subject-related components, demonstrating improved robustness in cross-subject and cross-session settings. Similarly, Zhang \textit{et al.}~\cite{zhang2025extended} introduced an extended variational autoencoder (VAE) framework that partitions the latent space into gesture-specific, subject-specific, and residual components to enhance generalization. Despite these advances, existing disentanglement-based approaches typically rely on architectural separation or reconstruction objectives and do not explicitly enforce a discriminative structure within the gesture-related latent representation. As a result, the learned features may still lack sufficient intra-class compactness or inter-class separability. Furthermore, many of these approaches depend on fixed weighting of multiple objectives, which can lead to unstable optimization when the relative scales of adversarial and discriminative losses differ significantly. This limitation highlights the need for adaptive mechanisms that balance competing objectives during training.


sEMG signals exhibit strong subject-dependent and session-dependent variability, posing a major challenge for reliable cross-subject gesture recognition. To address this issue, we propose a conservative subject-invariant learning framework that jointly optimizes multiple complementary objectives acting directly on the encoder representation. Here, \emph{conservative} denotes a training strategy that prevents any single objective from dominating the optimization process, thereby promoting stable and balanced multi-objective learning. Specifically, the framework integrates three components. First, an adversarial subject loss enforces subject invariance by encouraging the encoder, via a gradient-reversal mechanism, to remove subject-identifying information and align latent representations across users. Second, a gesture classification loss promotes intra-class compactness, ensuring that samples corresponding to the same gesture form cohesive clusters in the latent space despite variability in muscle activation, electrode displacement, or noise. Third, a triplet loss explicitly enforces inter-class separability by increasing the margin between embeddings of different gestures, thereby reducing overlap between gesture manifolds under subject or session shifts. To ensure stable optimization of these competing objectives, we further introduce a Lipschitz-inspired adaptive weighting mechanism that dynamically balances their contributions based on their relative scales. By combining structured representation learning with conservative, adaptive optimization, the proposed approach learns a latent representation that is simultaneously subject-invariant, discriminative, and well structured, providing a robust foundation for cross-subject sEMG-based gesture recognition. Unlike existing approaches, it explicitly addresses both representation structure and optimization stability within a unified framework.
The main contributions of this paper are summarized as follows:
\begin{itemize}
	\item We propose a conservative subject-invariant representation learning framework for sEMG-based gesture recognition that addresses inter-subject variability without requiring target-subject adaptation.
	
	\item We develop a multi-objective learning strategy that jointly integrates adversarial subject confusion, gesture classification, and triplet-based metric learning to enforce invariance, intra-class compactness, and inter-class separability.
	
	\item We introduce a Lipschitz-inspired adaptive weighting mechanism that dynamically balances competing objectives during training, improving optimization stability and reducing sensitivity to manual hyperparameter tuning.
	
	\item We demonstrate through extensive cross-subject evaluations that the proposed framework achieves more robust and consistent performance than existing adversarial and discriminative approaches.
\end{itemize}

The remainder of this paper is organized as follows. Section \ref{prop} provides the proposed conservative subject-invariant network (CSIN) methodology. Section \ref{res} presents the experimental setup and results. Section \ref{sec_Discussion} discusses future work. Finally, Section \ref{sec_Conclusion} concludes the paper. A list of notations used throughout the paper is provided in Table~\ref{tab:nomenclature}.

\begin{table}[!htp]
	\centering
	\caption{List of mathematical notation}
	\label{tab:nomenclature}
	\begin{tabular}{@{}ll@{}}
		\toprule
		Notation & Description \\
		\midrule
		
		$\mathbf{x}$ & Input feature vector extracted from EMG window \\
		$d$ & Dimensionality of input feature vector \\
		$\mathbf{z}$ & Latent representation (embedding) \\
		$p$ & Dimensionality of latent embedding \\
		
		$y_g$ & Gesture label \\
		$y_s$ & Subject label \\
		$C$ & Number of gesture classes \\
		$S$ & Number of subjects \\
		
		$f_{\theta}(\cdot)$ & Encoder network with parameters $\theta$ \\
		$h_{\phi}(\cdot)$ & Gesture classification head with parameters $\phi$ \\
		$h_{\psi}(\cdot)$ & Subject classification head with parameters $\psi$ \\
		
		$\hat{\mathbf{p}}_g$ & Predicted gesture class probabilities \\
		$\hat{\mathbf{p}}_s$ & Predicted subject class probabilities \\
		
		$\mathbf{W}_g, \mathbf{b}_g$ & Gesture head weight matrix and bias vector \\
		$\mathbf{W}_s, \mathbf{b}_s$ & Subject head weight matrix and bias vector \\
		
		$\mathcal{L}_{\mathrm{gest}}$ & Gesture classification loss \\
		$\mathcal{L}_{\mathrm{sub}}$ & Subject classification loss \\
		$\mathcal{L}_{\mathrm{adv}}$ & Adversarial loss for encoder \\
		$\mathcal{L}_{\mathrm{trip}}$ & Triplet metric learning loss \\
		$\mathcal{L}_{\mathrm{enc}}$ & Encoder optimization loss \\
		$\mathcal{L}_{\mathrm{total}}$ & Total training loss (for reporting) \\
		
		$\lambda_{\mathrm{gest}}$ & Gesture loss weight \\
		$\lambda_{\mathrm{adv}}$ & Adversarial loss weight \\
		$\lambda_{\mathrm{trip}}$ & Triplet loss weight \\
		$\lambda_{\mathrm{wd}}$ & Weight decay coefficient \\
		
		$\Theta$ & Set of all trainable parameters \\
		
		$r_{\mathrm{adv}}$ & Relative adversarial loss ratio \\
		$r_{\mathrm{trip}}$ & Relative triplet loss ratio \\
		$\hat{L}_{\mathrm{adv}}$ & Smoothed adversarial loss estimate \\
		$\hat{L}_{\mathrm{trip}}$ & Smoothed triplet loss estimate \\
		$\beta$ & Exponential smoothing coefficient \\
		$\epsilon$ & Small constant for numerical stability \\
		
		$\mathbf{z}^a_i$ & Anchor sample in triplet \\
		$\mathbf{z}^p_i$ & Positive sample (same gesture, different subject) \\
		$\mathbf{z}^n_i$ & Negative sample (different gesture) \\
		$d(\mathbf{u},\mathbf{v})$ & Squared Euclidean distance between $\mathbf{u}$ and $\mathbf{v}$ \\
		$\alpha$ & Margin parameter in triplet loss \\
		$N$ & Number of triplets in batch \\

		\bottomrule
	\end{tabular}
\end{table}

\section{Proposed method }\label{prop}

\begin{figure}[t]
	\centering
	\includegraphics[width=1\linewidth]{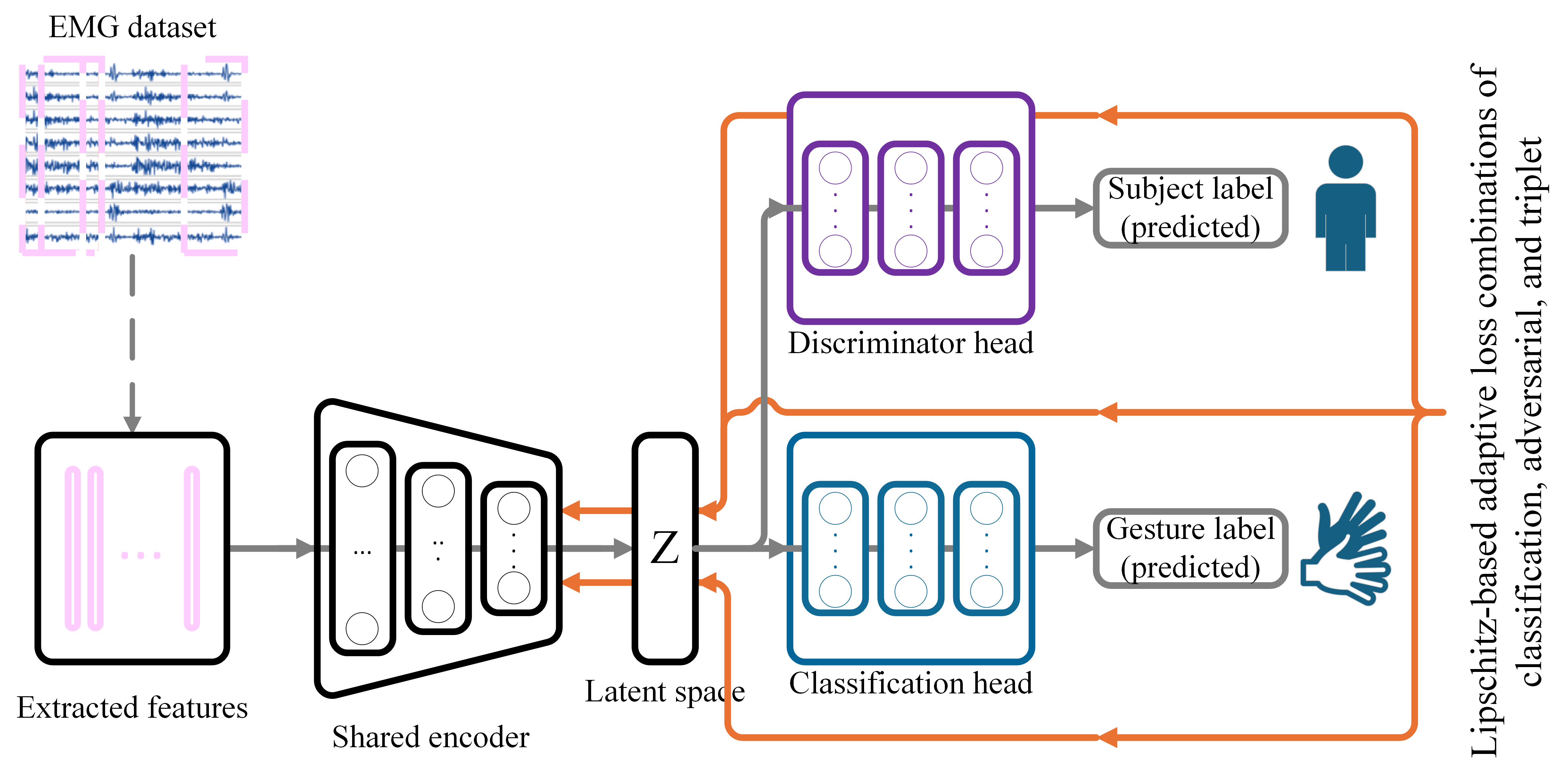} 
	\caption{Overview of the proposed conservative subject invariant network (CSIN) for sEMG-based gesture recognition. The model consists of a shared feature encoder and two heads: a gesture classification head and a subject classification head. The encoder learns a disentangled latent representation that is intra-class compact, inter-class separable, and invariant to subject-specific characteristics. Forward flow of data and backward gradients are shown in gray and orange solid lines, respectively. The feature extraction step is shown in a dashed line. }
	\label{fig:dsin_architecture}
\end{figure}

In this section, we present the proposed conservative subject-invariant network (CSIN) for gesture recognition from sEMG signals. The overall architecture is shown in Fig.~\ref{fig:dsin_architecture}. CSIN is built on a multi-head neural architecture with a shared feature encoder, designed to learn a latent representation invariant to subject-specific variations while preserving gesture-related factors. The method aims to obtain a representation that is simultaneously \emph{intra-class compact}, \emph{inter-class separable}, and \emph{subject invariant}. Intra-class compactness is encouraged by the gesture classification objective, which drives samples of the same gesture toward cohesive clusters in the latent space. Inter-class separability is enforced using a triplet loss, which increases the relative distance between embeddings of distinct gestures. To reduce subject-specific effects, we use an adversarial strategy: a subject classifier attempts to identify the subject, while the encoder, through gradient reversal, is trained to remove this information from the representation. To ensure stable and balanced optimization of these competing objectives, we further introduce a Lipschitz-inspired adaptive weighting strategy that dynamically adjusts the contributions of the adversarial and triplet losses with respect to the primary gesture-classification objective. By jointly optimizing these objectives with adaptive loss balancing, CSIN learns a representation that remains discriminative while being more robust to subject variability, leading to improved cross-subject generalization.

Let $\mathbf{x} \in \mathbb{R}^{d}$ denote the extracted feature vector from a window of sEMG signals, and let $y_g \in \{1,\dots, C\}$ and $y_s \in \{1,\dots, S\}$ denote the corresponding gesture and subject labels, respectively. A shared encoder is used to transform the input feature vector into a latent embedding:
\begin{equation}
	\mathbf{z} = f_{\theta}(\mathbf{x}) \in \mathbb{R}^{p},
\end{equation}
where $\theta$ denotes the encoder parameters and $p$ is the embedding dimension. The encoder maps the input to a latent embedding $\mathbf{z}$ that preserves gesture-discriminative information while suppressing subject-specific variations, forming the basis for learning representations that are both discriminative and invariant across subjects.

\subsection{Multi-head output structure}

The proposed architecture employs two primary prediction heads operating on the shared latent representation $\mathbf{z}$: a gesture classification head and a subject classification head. 

The gesture head is defined as:
\begin{equation}
	\hat{\mathbf{p}}_g = h_{\phi}(\mathbf{z}) = \mathrm{softmax}(\mathbf{W}_g \mathbf{z} + \mathbf{b}_g),
\end{equation}
where $\phi$ denotes the parameters of the gesture classifier. This head estimates the class probabilities over the $C$ gesture classes.

The subject head is defined as:
\begin{equation}
	\hat{\mathbf{p}}_s = h_{\psi}(\mathbf{z}) = \mathrm{softmax}(\mathbf{W}_s \mathbf{z} + \mathbf{b}_s),
\end{equation}
where $\psi$ denotes the parameters of the subject classifier. This head predicts the subject identity and is used exclusively for adversarial training.

During optimization, the gesture head encourages the encoder to learn gesture-discriminative representations. In contrast, the subject head is coupled with a gradient reversal layer, which inverts the gradients propagated to the encoder during backpropagation. As a result, the encoder is trained to suppress subject-specific information while retaining task-relevant features.

\subsection{Loss functions}

The proposed framework is formulated as a multi-objective optimization problem on the shared latent representation. Specifically, it jointly optimizes three objectives: (i) gesture classification, (ii) adversarial subject classification for invariance, and (iii) triplet-based metric regularization. 

\subsubsection*{1) Gesture Classification Loss}

The gesture head is trained using the standard cross-entropy loss as follows:
\begin{equation}
	\mathcal{L}_{\mathrm{gest}} = - \sum_{c=1}^{C} \mathbb{I}(y_g = c)\, \log \hat{p}_g(c),
\end{equation}
which encourages the encoder to learn gesture-discriminative representations. This term serves as the primary task objective and acts as the reference scale for balancing auxiliary losses during training.

\subsubsection*{2) Adversarial subject loss}

To suppress subject-dependent information in the latent representation, a subject classifier is trained jointly with the encoder. The subject head minimizes the cross-entropy loss as follows:
\begin{equation}
	\mathcal{L}_{\mathrm{sub}} = - \sum_{s=1}^{S} \mathbb{I}(y_s = s)\, \log \hat{p}_s(s).
\end{equation}

To enforce subject invariance, a gradient reversal mechanism is applied between the subject head and the encoder. During backpropagation, the gradients from the subject loss are multiplied by $-1$ before reaching the encoder. As a result, the subject head is trained to correctly predict subject identity, while the encoder is trained adversarially to remove subject-discriminative information. The adversarial contribution to the encoder objective is therefore given by $\mathcal{L}_{\mathrm{adv}} = - \mathcal{L}_{\mathrm{sub}}$. This adversarial interaction encourages the latent representation to be invariant to subject identity while retaining task-relevant gesture information.

\subsubsection*{3) Triplet loss}

To enhance the discriminative structure of the latent space, we employ a triplet loss that enforces both intra-class compactness and inter-class separation. Each training sample is organized into triplets of the form $(\mathbf{z}^a_i,\mathbf{z}^p_i,\mathbf{z}^n_i)$, where the anchor and positive embeddings correspond to the same gesture class (typically from different subjects). In contrast, the negative embedding belongs to a different gesture class.

Using the squared Euclidean distance as follows:
\begin{equation}
	d(\mathbf{u},\mathbf{v}) = \|\mathbf{u}-\mathbf{v}\|_2^2,
\end{equation}
the triplet loss is defined as:
\begin{equation}
	\mathcal{L}_{\mathrm{trip}} = \frac{1}{N}\sum_{i=1}^{N} \left[ d(\mathbf{z}^a_i,\mathbf{z}^p_i)-d(\mathbf{z}^a_i,\mathbf{z}^n_i)+\alpha\right]_+,
\end{equation}
where $\alpha > 0$ is a margin parameter and $[\cdot]_+ = \max(0,\cdot)$. This objective brings embeddings of the same gesture closer together while pushing those of different gestures further apart, improving class separability and reducing overlap in the latent space.

\subsection{Lipschitz-adaptive multi-objective optimization}

A key challenge in joint optimization is the relative scaling of the adversarial and triplet objectives with respect to the main gesture-classification loss. Fixed coefficients often require extensive manual tuning and may lead to unstable training when one auxiliary objective dominates the others. To address this issue, we introduce a Lipschitz-inspired adaptive weighting strategy that dynamically adjusts the weights of the adversarial and triplet losses during optimization.

The total loss used is written as
\begin{equation}
	\mathcal{L}_{\mathrm{total}}=\lambda_{\mathrm{gest}} \mathcal{L}_{\mathrm{gest}}+\lambda_{\mathrm{adv}} \mathcal{L}_{\mathrm{sub}}+\lambda_{\mathrm{trip}} \mathcal{L}_{\mathrm{trip}}+\lambda_{\mathrm{wd}} \|\Theta\|_2^2,
\end{equation}
where $\Theta$ collects all trainable weights and $\lambda_{\mathrm{wd}}$ denotes the weight-decay coefficient. Since the encoder is trained adversarially with respect to subject identity, the loss used to compute gradients for the shared encoder becomes
\begin{equation}\label{eq:encoder_loss_placeholder}
	\mathcal{L}_{\mathrm{enc}}=\lambda_{\mathrm{gest}} \mathcal{L}_{\mathrm{gest}}-\lambda_{\mathrm{adv}} \mathcal{L}_{\mathrm{sub}}+\lambda_{\mathrm{trip}} \mathcal{L}_{\mathrm{trip}}+\lambda_{\mathrm{wd}} \|\Theta\|_2^2.
\end{equation}

In contrast, the subject-classification head itself is trained in the standard direction to minimize $\mathcal{L}_{\mathrm{sub}}$. This results in the usual adversarial interaction between the encoder and the subject classifier.

To avoid manually fixing $\lambda_{\mathrm{adv}}$ and $\lambda_{\mathrm{trip}}$, we estimate their relative magnitudes with respect to the gesture loss. To simplify the notation, we treat the subject classification loss $\mathcal{L}_{\mathrm{sub}}$ as a proxy for the adversarial loss magnitude, as both operate on a similar scale. Let
\begin{equation}
	r_{\mathrm{adv}} = \frac{\mathcal{L}_{\mathrm{sub}}}{\mathcal{L}_{\mathrm{gest}} + \epsilon}, 
	\qquad
	r_{\mathrm{trip}} = \frac{\mathcal{L}_{\mathrm{trip}}}{\mathcal{L}_{\mathrm{gest}} + \epsilon},
\end{equation}
computed at predefined intervals during training, whenever $\mathcal{L}_{\mathrm{gest}}$ is sufficiently large. These ratios act as practical proxies for the relative Lipschitz scales of the objectives, since the magnitude of a loss is related to the norm of its gradients during stochastic optimization. We then maintain exponentially smoothed estimates as follows:
\begin{equation}
	\hat{L}_{\mathrm{adv}}^{(t)}=\beta \hat{L}_{\mathrm{adv}}^{(t-1)}+(1-\beta)\, r_{\mathrm{adv}}^{(t)},
\end{equation}

\begin{equation}
	\hat{L}_{\mathrm{trip}}^{(t)}=\beta \hat{L}_{\mathrm{trip}}^{(t-1)}+(1-\beta)\, r_{\mathrm{trip}}^{(t)},
\end{equation}
where $\beta \in [0,1)$ is a smoothing coefficient.

The adaptive weights are then updated inversely proportional to these estimated scales,
\begin{equation}
	\lambda_{\mathrm{adv}}^{(t)} = \frac{\lambda_{\mathrm{gest}}}{\hat{L}_{\mathrm{adv}}^{(t)}},
	\qquad
	\lambda_{\mathrm{trip}}^{(t)} = \frac{\lambda_{\mathrm{gest}}}{\hat{L}_{\mathrm{trip}}^{(t)}}.
\end{equation}
This update rule ensures that auxiliary objectives remain on a scale comparable to the main classification objective, thereby improving optimization stability and reducing the need for exhaustive hyperparameter tuning.

To improve stability, the adaptive updates are applied only after an initial warm-up phase. The resulting weights are also clipped to predefined ranges,
\begin{equation}
	\lambda_{\mathrm{adv}} \in [\lambda_{\mathrm{adv}}^{\min}, \lambda_{\mathrm{adv}}^{\max}],
	\qquad
	\lambda_{\mathrm{trip}} \in [\lambda_{\mathrm{trip}}^{\min}, \lambda_{\mathrm{trip}}^{\max}],
\end{equation}
to avoid extreme values caused by noisy mini-batch estimates.

This Lipschitz-adaptive weighting strategy provides two main advantages. First, it prevents the adversarial and metric-learning terms from dominating the gesture objective. Second, it reduces the sensitivity of the training process to dataset-specific tuning, which is particularly important in cross-subject EMG learning, where the balance between invariance and discrimination can vary across datasets.

\paragraph{Why Lipschitz-Adaptive Balancing Improves Training Stability}

The proposed adaptive weighting strategy can be interpreted through the lens of Lipschitz continuity and gradient scaling. In multi-objective optimization, each loss term contributes gradients of potentially different magnitudes with respect to the shared encoder parameters. When fixed coefficients are used, the optimization process becomes highly sensitive to the relative scales of the gradients, often leading to one objective dominating the others or to unstable oscillatory behavior.

By estimating the relative magnitude of each auxiliary loss with respect to the primary gesture-classification loss, the proposed method effectively balances their gradient contributions. Since the norm of a loss gradient is related to its Lipschitz constant, the ratio $\frac{\mathcal{L}_{\mathrm{sub}}}{\mathcal{L}_{\mathrm{gest}}}$ and $\frac{\mathcal{L}_{\mathrm{trip}}}{\mathcal{L}_{\mathrm{gest}}}$ serves as a practical indicator of the relative scale of each objective. Adjusting the weights inversely to these ratios therefore acts as a form of gradient normalization, helping prevent any single objective from dominating the shared representation.

Moreover, exponential smoothing helps stabilize these estimates across mini-batches, reducing the impact of stochastic fluctuations and avoiding abrupt changes during training. Together, inverse scaling and temporal smoothing lead to a more stable and adaptive optimization process, making the model less sensitive to initialization and dataset-specific variations. From a representation learning perspective, this results in a better balance between discriminability (gesture classification), invariance (adversarial subject confusion), and structural organization (triplet loss), producing embeddings that generalize more effectively to unseen subjects. This balance is especially important in cross-subject EMG recognition, where the relative difficulty of these objectives can vary across individuals.

\subsection{Optimization details}

The network is trained using the Adam optimizer. For each mini-batch, the gesture, subject, triplet, and reconstruction losses are computed from the shared embedding. The encoder is then updated using the adversarially signed objective in \eqref{eq:encoder_loss_placeholder}, while the subject head is updated in the normal direction using subject classification loss.

Gradient clipping is used to avoid exploding updates, and $L_2$ regularization is applied to the trainable weights. During inference, only the encoder and gesture-classification head are retained; the subject-classification and reconstruction heads are used only during training.

\par The complete training procedure of the proposed framework is summarized in Algorithm~\ref{alg:csin_main}. For each mini-batch, the encoder produces a latent representation that is simultaneously optimized for gesture discrimination, subject invariance, and metric structure. The gesture loss acts as the primary task objective, while the adversarial subject loss and triplet loss are adaptively reweighted during training using the proposed Lipschitz-inspired balancing mechanism. After an initial warm-up stage, the relative magnitudes of the subject and triplet losses are periodically estimated relative to the gesture loss, and their coefficients are updated via exponentially smoothed inverse scaling. This allows the network to maintain stable multi-objective optimization without relying on manually fixed loss weights. During inference, only the encoder and gesture head are used.

\begin{algorithm}[!t]
	\caption{Lipschitz-Adaptive Training Procedure}
	\label{alg:csin_main}
	\small
	\KwIn{Training set $\mathcal{D}=\{(\mathbf{x}_i,y_g^i,y_s^i)\}_{i=1}^{N}$}
	\KwIn{Initial weights $\lambda_{\mathrm{gest}}, \lambda_{\mathrm{adv}}, \lambda_{\mathrm{trip}}$}
	\KwIn{Margin $\alpha$, smoothing coefficient $\beta$, warm-up length $T_{\mathrm{warm}}$, update interval $K$}
	\KwOut{Trained encoder $f_\theta$ and gesture head $h_{\phi_g}$}
	
	Initialize encoder, gesture head, subject head, and reconstruction head parameters\;
	Initialize optimizer state\;
	Initialize $\hat{L}_{\mathrm{adv}} \leftarrow 1$, $\hat{L}_{\mathrm{trip}} \leftarrow 1$\;
	
	\For{epoch $=1$ \KwTo $T_{\max}$}{
		Shuffle training data\;
		
		\For{each mini-batch $\mathcal{B}$}{
			Compute latent embeddings $\mathbf{z}=f_\theta(\mathbf{x})$\;
			Compute gesture predictions $\hat{\mathbf{p}}_g=h_{\phi_g}(\mathbf{z})$\;
			Compute subject predictions $\hat{\mathbf{p}}_s=h_{\phi_s}(\mathbf{z})$\;
			
			Compute $\mathcal{L}_{\mathrm{gest}}$, $\mathcal{L}_{\mathrm{sub}}$, and $\mathcal{L}_{\mathrm{trip}}$ \;
			
			Form encoder loss:
			\[
			\mathcal{L}_{\mathrm{enc}}
			=
			\lambda_{\mathrm{gest}}\mathcal{L}_{\mathrm{gest}}
			-
			\lambda_{\mathrm{adv}}\mathcal{L}_{\mathrm{sub}}
			+
			\lambda_{\mathrm{trip}}\mathcal{L}_{\mathrm{trip}}.
			\]
			
			Update encoder and auxiliary heads using backpropagation\;
			
			\If{training step $> T_{\mathrm{warm}}$ and update step is a multiple of $K$}{
				Compute
				\[
				r_{\mathrm{adv}}=\frac{\mathcal{L}_{\mathrm{sub}}}{\mathcal{L}_{\mathrm{gest}}},
				\qquad
				r_{\mathrm{trip}}=\frac{\mathcal{L}_{\mathrm{trip}}}{\mathcal{L}_{\mathrm{gest}}}
				\]
				when $\mathcal{L}_{\mathrm{gest}}>\epsilon$\;
				
				Update smoothed estimates
				\[
				\hat{L}_{\mathrm{adv}} \leftarrow \beta \hat{L}_{\mathrm{adv}} + (1-\beta)r_{\mathrm{adv}},
				\]
				\[
				\hat{L}_{\mathrm{trip}} \leftarrow \beta \hat{L}_{\mathrm{trip}} + (1-\beta)r_{\mathrm{trip}}.
				\]
				
				Update adaptive coefficients
				\[
				\lambda_{\mathrm{adv}} \leftarrow \frac{\lambda_{\mathrm{gest}}}{\hat{L}_{\mathrm{adv}}},
				\qquad
				\lambda_{\mathrm{trip}} \leftarrow \frac{\lambda_{\mathrm{gest}}}{\hat{L}_{\mathrm{trip}}}.
				\]
				
				Clip $\lambda_{\mathrm{adv}}$ and $\lambda_{\mathrm{trip}}$ to predefined ranges\;
			}
		}
	}
\end{algorithm}

\section{Results}\label{res}
\subsection{Datasets}

\subsubsection{UCI EMG dataset}
The UCI dataset\footnote{\url{https://archive.ics.uci.edu/dataset/481/emg+data+for+gestures}} comprises raw EMG recordings captured using a MYO Thalmic bracelet positioned on the forearm of 36 healthy subjects. During data collection, participants executed a series of static hand gestures. At the same time, eight evenly distributed sensors around the forearm recorded myographic signals at a sampling rate, yielding approximately 40,000-50,000 time-stamped measurements per recording session. Each participant completed two separate trials consisting of six to seven basic hand movements (here we use 6 gestures), with each gesture maintained for 3 seconds, followed by a 3-second rest period. The recorded gestures included resting hand position, fist clenching, wrist flexion and extension, radial and ulnar deviations, and an extended palm configuration (though not all subjects performed this). Windows of length 250ms with a 50ms step length are used for feature extraction. Four features of mean absolute value (MAV), root mean square (RMS), wave length (WL), zero crossing (ZC) are extracted.

\subsubsection{NinaPro DB5 dataset}
The Ninapro DB5 dataset\footnote{\url{https://ninapro.hevs.ch/instructions/DB5.html}} includes surface EMG recordings from 10 intact subjects performing hand and wrist movements \cite{pizzolato2017comparison}. Data were acquired using two Thalmic Myo Armbands, each equipped with 8 dry electrodes equally spaced around the forearm, providing a total of 16 EMG channels sampled at 200 Hz. The first armband was positioned at the height of the radio-humeral joint, while the second was placed with a 22.5-degree clockwise rotation. For this study, we utilize only the EMG channels and focus on a subset of 10 movement classes: Rest, Finger Abduction, Fist, Finger Adduction, Middle Axis Supination, Middle Axis Pronation, Wrist Flexion, Wrist Extension, Radial Deviation, and Ulnar Deviation. These movements span basic finger articulations and fundamental wrist motions, representing functionally relevant gestures for prosthetic control and human-computer interaction applications. Each movement repetition lasted approximately 5 seconds, followed by a 3-second rest, for a total of 6 repetitions per gesture class. The movements were organized into three exercises: basic finger movements; isometric and isotonic hand configurations with wrist movements; and functional grasping movements. The dataset provides synchronized recordings, including raw EMG signals and movement labels with refined temporal annotations (restimulus) that more accurately reflect the actual movement execution. The same features as in the UCI dataset are extracted from standard windows of length 250ms with a 50ms step size for suitable real-time use \cite{smith2010determining}.

\subsection{Network architecture}

\par The proposed CSIN network consists of a shared encoder followed by two specialized prediction heads for gesture classification and subject classification. The complete architecture is detailed in Table \ref{tab:network_architecture}. The shared encoder processes z-score normalized 32/64-dimensional input features (8/16 electrodes $\times$ 4 features: MAV, RMS, WL, ZC) through two fully connected blocks of 64 units each, with layer normalization, ReLU activation, and 30\% dropout. The encoder outputs a 16-dimensional embedding that serves as the bottleneck representation for both downstream heads. The gesture classification head consists of three 16-unit fully connected layers with ReLU activations, layer normalization, and dropout, followed by a 6/10-way softmax output for the UCI/NinaPro dataset. The subject classification head mirrors this structure but uses 36-unit layers and outputs 35/10 classes (corresponding to training subjects in leave-one-subject-out cross-validation). During training, the subject head minimizes cross-entropy loss with normal gradients, while the shared encoder receives reversed gradients from this head to remove subject-identifying information. Training employs the Adam optimizer ($\beta_1=0.9$, $\beta_2=0.999$, $\epsilon=10^{-8}$) with a fixed learning rate of 0.001, $L2$ regularization ($10^{-4}$), batch size of 512, and 300 epochs. The total loss combines gesture classification ($\lambda_{gest}$), adversarial subject loss ($\lambda_adv$), and triplet loss ($\lambda_{trip}$, margin $\alpha=0.5$). Triplets are constructed with anchor and positive samples sharing the same gesture from different subjects, while negative samples represent different gestures, enforcing both inter-class separability and cross-subject consistency.

\begin{table}[t]
	\centering
	\caption{CSIN Network Architecture for the UCI/NinaPro dataset.}
	\label{tab:network_architecture}
	\begin{tabular}{@{}lll@{}}
		\toprule
		\textbf{Component} & \textbf{Configuration} & \textbf{Output} \\ \midrule
		Input & Z-score normalization & 32/64 \\ \midrule
		\multicolumn{3}{l}{\textit{Shared Encoder}} \\
		\quad FC Block 1 & FC(64) + LN + ReLU + Drop(0.3) & 64 \\
		\quad FC Block 2 & FC(64) + LN + ReLU + Drop(0.3) & 64 \\
		\quad Embedding & FC(16) & 16 \\ \midrule
		\multicolumn{3}{l}{\textit{Gesture Head}} \\
		\quad FC Blocks & 3$\times$[ReLU + LN + Drop + FC(16)] & 16 \\
		\quad Output & FC(6/10) + Softmax & 6/10 \\ \midrule
		\multicolumn{3}{l}{\textit{Subject Head}} \\
		\quad FC Blocks & 3$\times$[ReLU+LN+Drop+FC(36)] & 36 \\
		\quad Output & FC(35/10) + Softmax & 35/10 \\ \bottomrule
	\end{tabular}
\end{table}

\subsection{Ablation study on training schedules}
\label{sec:ablation_scenarios}

In addition to the main Lipschitz-adaptive training strategy, we investigated whether the timing of adversarial interaction between the encoder and the subject-classification head affects final performance. In all cases, the network architecture, feature extraction procedure, and loss definitions remained unchanged; only the update schedule of the encoder and subject head differed.

We considered four training schedules:

\paragraph{Scenario 1 (fully coupled training)}
All modules are trained jointly from the first epoch onward. The encoder is optimized with the full objective, while the subject head is trained continuously with a subject-classification loss.

\paragraph{Scenario 2 (warm-up without subject-head training)}
During the initial stage, the subject head is frozen, and the encoder is optimized only with gesture and triplet losses. After the warm-up period, adversarial subject learning is activated, and full joint training begins.

\paragraph{Scenario 3 (delayed adversarial interaction)}
The subject head is trained from the start, but the encoder does not receive adversarial subject gradients during the warm-up stage. After initialization, the full objective is used for encoder training.

\paragraph{Scenario 4 (early subject supervision with later freezing)}
The subject head is trained during the warm-up stage to establish a discriminative subject signal. After initialization, the subject head is frozen, while the encoder continues to receive adversarial gradients from the fixed subject classifier.

\begin{table}[!t]
	\centering
	\caption{Training-schedule ablation scenarios. Check marks indicate active losses or updated modules.}
	\label{tab:scenario_summary}
	\renewcommand{\arraystretch}{1.2}
	\setlength{\tabcolsep}{4pt}
	\begin{adjustbox}{width=\linewidth}
		\begin{tabular}{p{1.2cm} p{1.8cm} p{3.4cm} p{2.1cm} p{2.1cm}}
			\hline
			\textbf{Scenario} & \textbf{Stage} & \textbf{Encoder objective} & \textbf{Gesture head} & \textbf{Subject head} \\
			\hline
			S1 & Full training &
			$\mathcal{L}_{\mathrm{gest}} - \lambda_{\mathrm{adv}}\mathcal{L}_{\mathrm{sub}} + \lambda_{\mathrm{trip}}\mathcal{L}_{\mathrm{trip}}$
			& updated & updated \\
			\hline
			
			S2 & Warm-up &
			$\mathcal{L}_{\mathrm{gest}} + \lambda_{\mathrm{trip}}\mathcal{L}_{\mathrm{trip}}$
			& updated & frozen \\
			& Later stage &
			$\mathcal{L}_{\mathrm{gest}} - \lambda_{\mathrm{adv}}\mathcal{L}_{\mathrm{sub}} + \lambda_{\mathrm{trip}}\mathcal{L}_{\mathrm{trip}}$
			& updated & updated \\
			\hline
			
			S3 & Warm-up &
			$\mathcal{L}_{\mathrm{gest}} + \lambda_{\mathrm{trip}}\mathcal{L}_{\mathrm{trip}}$
			& updated & updated \\
			& Later stage &
			$\mathcal{L}_{\mathrm{gest}} - \lambda_{\mathrm{adv}}\mathcal{L}_{\mathrm{sub}} + \lambda_{\mathrm{trip}}\mathcal{L}_{\mathrm{trip}}$
			& updated & updated \\
			\hline
			
			S4 & Warm-up &
			$\mathcal{L}_{\mathrm{gest}} + \lambda_{\mathrm{trip}}\mathcal{L}_{\mathrm{trip}}$
			& updated & updated \\
			& Later stage &
			$\mathcal{L}_{\mathrm{gest}} - \lambda_{\mathrm{adv}}\mathcal{L}_{\mathrm{sub}} + \lambda_{\mathrm{trip}}\mathcal{L}_{\mathrm{trip}}$
			& updated & frozen \\
			\hline
		\end{tabular}
	\end{adjustbox}
\end{table}

Table~\ref{tab:scenario_summary} summarizes these schedules. The purpose of this ablation is to determine whether the final performance gains arise primarily from the proposed Lipschitz-adaptive balancing itself or from a particular adversarial training schedule. As shown in the experimental results, the differences among scenarios are relatively small, indicating that the proposed method is robust to the exact scheduling strategy. This observation suggests that the main benefit derives from the multi-objective, subject-invariant formulation together with adaptive loss balancing, rather than from a highly specialized training schedule.

\begin{table}[!th]
	\caption{Comparing the accuracies' average, standard deviation, maximum, and minimum (\%) of the proposed method with SOTA for the UCI EMG dataset. Best results are bold.}
	\label{tab1}
	\begin{adjustbox}{width=\linewidth}
		\begin{tabular}{@{}lcccccc@{}}
			\toprule
			Method   &                Mean $\uparrow$                                                                                                                                                                                                                                                                                                                                                                                                                                                                                                                                                                                                                                          &         Std $\downarrow$         & Max $\uparrow$ & Min $\uparrow$&Window size(ms)                                                                                                                                                                                                                                                                                                                                                                                                                                                                                                                                                                                                                                                & Number of gestures                                                                                                                                                                                                                                                                                                                                                                                                                                                                                                                                                                                                                                                 \\ \midrule
			KNN-1 \cite{lee2024decoding}                                                                                            &       77.5                                                                                                                                                                                                                                                                                                                                                                                                                                                                                                                                                                                                                &                    -    & -&- & 300                                                                                                                                                                                                                                                                                                                                                                                                                                                                                                                                                                                                & 7                                                                                                                                                                                                                                                                                                                                                                                                                                                                                                                                                                                                
			\\
			EMGBench \cite{yang2024emgbench}                                                                                            &            78.2                                                                                                                                                                                                                                                                                                                                                                                                                                                                                                                                                                                                           &                 -       &-  &-& 250                                                                                                                                                                                                                                                                                                                                                                                                                                                                                                                                                                                                & 6                                                                                                                                                                                                                                                                                                                                                                                                                                                                                                                                                                                                  
			\\\midrule
			Na\"{\i}ve CSIN&80.88&\textbf{15.05}&\textbf{98.12}&\textbf{32.93}&250& 6
			\\
			Full CSIN&\textbf{84.48}&15.74&97.49&22.36&250&6
			\\\bottomrule
		\end{tabular}
	\end{adjustbox}
\end{table}

\begin{table}[!th]
	\caption{Comparing the accuracies' average, standard deviation, maximum, and minimum (\%) of the proposed method with SOTA for the NinaPro DB5 EMG dataset. Best results are bold.}
	\label{tab2}
	\begin{adjustbox}{width=\linewidth}
		\begin{tabular}{@{}lcccccc@{}}
			\toprule
			Method                                                                &                Mean $\uparrow$        &                    Std $\downarrow$ & Max $\uparrow$ & Min $\uparrow$  & Window size (ms)                                                                                                                                                                                                                                                                                                                                                                                                                                                                                                                                                                                                                                                & Number of gestures \\ \midrule
			FedEMG \cite{lee2025fedemg}                                                                                             &            24.52            &       -   &           -   & - & 1000                                                                                                                                                                                                                                                                                                                                                                                                                                                                                                                                                                                                & 53                                                                                                                                                                                                                                                                                                                                                                                                                                                                                                                                                                                                
			\\
			KNN-1 \cite{lee2024decoding}                                                                                             &        68.7                &    -   &        -         &-  & 500                                                                                                                                                                                                                                                                                                                                                                                                                                                                                                                                                                                                & 53                                                                                                                                                                                                                                                                                                                                                                                                                                                                                                                                                                                                 \\ 
			genANN \cite{hoshino2024comparison}                                                                                             &       17.28                 &     4.82  &           -      & - & 250                                                                                                                                                                                                                                                                                                                                                                                                                                                                                                                                                                                                & 40                                                                                                                                                                                                                                                                                                                                                                                                                                                                                                                                                                                                 \\ 
			EMGBench \cite{yang2024emgbench}                                                                                             &         41.30               &      -    &       -       & - & 250                                                                                                                                                                                                                                                                                                                                                                                                                                                                                                                                                                                                & 10                                                                                                                                                                                                                                                                                                                                                                                                                                                                                                                                                                                                  \\ \midrule
			Na\"{\i}ve CSIN&60.77&6.06&\textbf{65.52}&46.00&250& 10
			\\
			Full CSIN &\textbf{61.44}&\textbf{1.23}&62.79&\textbf{59.11}&250& 10
			
			\\
			\bottomrule
		\end{tabular}
	\end{adjustbox}
\end{table}

\begin{table*}[!th]
	\caption{Comparing the accuracies' average, standard deviation, maximum, and minimum (\%) of the proposed method with regular training for the UCI EMG dataset. Best and second-best results are bold and underlined, respectively.}
	\label{tab11}
	\begin{adjustbox}{width=\textwidth}
		\begin{tabular}{@{}lcccccccccccccccccc@{}}
			\toprule
			Methods                & \multicolumn{1}{l}{$\lambda_{\mathrm{adv}}$} & \multicolumn{1}{l}{$\lambda_{\mathrm{trip}}$} & \multicolumn{4}{c}{Scenario 1}                                                                                                                       & \multicolumn{4}{c}{Scenario 2}                                                                                                                       & \multicolumn{4}{c}{Scenario 3}                                                                                                                       & \multicolumn{4}{c}{Scenario 4}                                                                                                                       \\ \hline
			&                         &                         & \multicolumn{1}{l}{Mean $\uparrow$} & \multicolumn{1}{l}{Std $\downarrow$} & \multicolumn{1}{l}{Max $\uparrow$} & \multicolumn{1}{l}{Min $\uparrow$} & \multicolumn{1}{l}{Mean $\uparrow$} & \multicolumn{1}{l}{Std $\downarrow$} & \multicolumn{1}{l}{Max $\uparrow$} & \multicolumn{1}{l}{Min $\uparrow$} & \multicolumn{1}{l}{Mean $\uparrow$} & \multicolumn{1}{l}{Std $\downarrow$} & \multicolumn{1}{l}{Max $\uparrow$} & \multicolumn{1}{l}{Min $\uparrow$} & \multicolumn{1}{l}{Mean $\uparrow$} & \multicolumn{1}{l}{Std $\downarrow$} & \multicolumn{1}{l}{Max $\uparrow$} & \multicolumn{1}{l}{Min $\uparrow$} \\\cline{4-19} 
			Baseline                                                                                                                                     & 0                       & 0                       & \multicolumn{1}{l}{81.75}           & \multicolumn{1}{l}{14.33}            & \multicolumn{1}{l}{98.35}          & \multicolumn{1}{l}{36.71}          & \multicolumn{1}{l}{81.75}           & \multicolumn{1}{l}{14.33}            & \multicolumn{1}{l}{98.35}          & \multicolumn{1}{l}{36.71}          & \multicolumn{1}{l}{81.75}           & \multicolumn{1}{l}{14.33}            & \multicolumn{1}{l}{98.35}          & \multicolumn{1}{l}{36.71}          & \multicolumn{1}{l}{81.75}           & \multicolumn{1}{l}{14.33}            & \multicolumn{1}{l}{98.35}          & \multicolumn{1}{l}{36.71}          \\ \hline
			\multirow{13}{*}{\rotatebox{90}{\begin{tabular}[c]{@{}l@{}}CSIN with fixed weights\\ (gesture, adversarial, \\and triplet weights)\end{tabular}}} & 0                       & 0.1                     & 81.69                               & 15.09                                & 98.26                              & 33.84                              & 81.69                               & 15.09                                & 98.26                              & 33.84                              & 81.69                               & 15.09                                & 98.26                              & 33.84                              & 81.69                               & 15.09                                & 98.26                              & 33.84                              \\ 
			&                         & 1                       & 81.41                               & 14.12                                & 97.87                              & 36.71                              & 81.41                               & 14.12                                & 97.87                              & 36.71                              & 81.41                               & 14.12                                & 97.87                              & 36.71                              & 81.41                               & 14.12                                & 97.87                              & 36.71                              \\
			&                         & 10                      & 81.77                               & 14.22                                & 97.76                              & 36.71                              & 81.77                               & 14.22                                & 97.76                              & 36.71                              & 81.77                               & 14.22                                & 97.76                              & 36.71                              & 81.77                               & 14.22                                & 97.76                              & 36.71                              \\ \cline{2-19} 
			& \multirow{5}{*}{0.1}    & 0                       & 81.68                               & 14.14                                & 96.71                              & 38.22                              & 81.32                               & 14.00                                   & 97.81                              & 38.07                              & 81.42                               & 14.51                                & 97.96                              & 38.37                              & 81.71                               & 14.86                                & 98.40                               & 32.33                              \\
			&                         & 0.01                    &             81.74                        & 14.50                                     &    97.99                                &        38.82                            &           81.41                          &                     14.62                 &                                   97.76 &              35.95                      &               81.40                      &      14.24                                &     98.43                               &         37.61                           &           81.51                          &            14.87                          &                      98.40              &                    33.69                \\
			&                         & 0.1                     & 81.54                               & 14.37                                & 97.52                              & 32.63                              & 81.83                               & 14.30                                 & 97.87                              & 36.71                              & 81.46                               & 15.08                                & 98.28                              & 33.69                              & 81.51                               & 14.87                                & 98.40                               & 33.69                              \\
			&                         & 1                       & 81.87                               & 14.21                                & 97.76                              & 36.70                               & 81.68                               & 15.04                                & 98.23                              & 33.38                              & 81.46                               & 15.08                                & 98.28                              & 33.69                              & 81.51                               & 14.87                                & 98.4                               & 33.69                              \\
			&                         & 10                      & 81.74                               & 14.59                                & 97.99                              & 38.82                              & 81.69                               & \textbf{13.90}                                 & 97.34                              & \underline{40.63}                              & 81.40                                & 14.58                                & \textbf{98.75}                              & 33.38                              & \underline{81.92}                               & 14.40                                 & \textbf{98.75}                              & 38.82                              \\ \cline{2-19} 
			& \multirow{5}{*}{0.01}   & 0                       & 81.50                                & 14.66                                & \underline{98.47}                              & 33.23                              & 81.68                               & 15.04                                & 98.23                              & 33.38                              & 81.46                               & 15.08                                & 92.28                              & 33.69                              & 81.31                               & 14.91                                & 98.40                               & 33.69                              \\
			&                         & 0.01                    & 81.32                               & 14.91                                & 98.11                              & 31.72                              & 81.83                               & 14.30                                 & 97.87                              & 36.71                              & 81.44                               & 14.90                                 & 98.43                              & 34.74                              & 81.51                               & 14.87                                & 98.40                               & 33.69                              \\
			&                         & 0.1                     & 81.63                               & 14.10                                 & 97.18                              & 32.93                              & 98.70                                & 15.06                                & 98.70                               & 30.82                              & 81.46                               & 15.08                                & 98.28                              & 33.69                              & 81.91                               & 14.66                                & 98.40                               & 30.51                              \\
			&                         & 1                       & 81.68                               & 14.14                                & 96.71                              & 38.22                              & 81.80                                & 14.11                                & 98.11                              & \textbf{41.08}                              & 81.51                               & 14.59                                & 97.73                              & 38.22                              & \textbf{82.26}                               & \underline{13.99}                                & 98.23                              & 37.76                              \\
			&                         & 10                      & 81.62                               & 14.42                                & 98.43                              & 35.35                              & 81.83                               & 14.30                                 & 97.87                              & 36.71                              & 81.44                               & 14.90                                 & 98.43                              & 34.74                              & 81.51                               & 14.87                                & 98.40                               & 33.69                              \\ \bottomrule
		\end{tabular}
	\end{adjustbox}
\end{table*}

\begin{table*}[!th]
	\caption{Comparing the accuracies' average, standard deviation, maximum, and minimum (\%) of the proposed method with regular training for the NinaPro DB5 EMG dataset. Best and second-best results are bold and underlined, respectively.}
	\label{tab22}
	\begin{adjustbox}{width=\textwidth}
		\begin{tabular}{@{}lcccccccccccccccccc@{}}
			\toprule
			Methods   & \multicolumn{1}{l}{$\lambda_{\mathrm{adv}}$} & \multicolumn{1}{l}{$\lambda_{\mathrm{trip}}$} & \multicolumn{4}{c}{Scenario 1}                                                                                                                       & \multicolumn{4}{c}{Scenario 2}                                                                                                                       & \multicolumn{4}{c}{Scenario 3}                                                                                                                       & \multicolumn{4}{c}{Scenario 4}                                                                                                                       \\ \hline
			&                         &                         & \multicolumn{1}{l}{Mean $\uparrow$} & \multicolumn{1}{l}{Std $\downarrow$} & \multicolumn{1}{l}{Max $\uparrow$} & \multicolumn{1}{l}{Min $\uparrow$} & \multicolumn{1}{l}{Mean $\uparrow$} & \multicolumn{1}{l}{Std $\downarrow$} & \multicolumn{1}{l}{Max $\uparrow$} & \multicolumn{1}{l}{Min $\uparrow$} & \multicolumn{1}{l}{Mean $\uparrow$} & \multicolumn{1}{l}{Std $\downarrow$} & \multicolumn{1}{l}{Max $\uparrow$} & \multicolumn{1}{l}{Min $\uparrow$} & \multicolumn{1}{l}{Mean $\uparrow$} & \multicolumn{1}{l}{Std $\downarrow$} & \multicolumn{1}{l}{Max $\uparrow$} & \multicolumn{1}{l}{Min $\uparrow$} \\\cline{4-19} 
			Baseline                                                                                                                                     & 0                       & 0                       & \multicolumn{1}{l}{60.85}           & \multicolumn{1}{l}{5.60}            & \multicolumn{1}{l}{65.49}          & \multicolumn{1}{l}{47.11}          & \multicolumn{1}{l}{60.85}           & \multicolumn{1}{l}{5.60}            & \multicolumn{1}{l}{65.49}          & \multicolumn{1}{l}{47.11}          & \multicolumn{1}{l}{60.85}           & \multicolumn{1}{l}{5.60}            & \multicolumn{1}{l}{65.49}          & \multicolumn{1}{l}{47.11}          & \multicolumn{1}{l}{60.85}           & \multicolumn{1}{l}{5.60}            & \multicolumn{1}{l}{65.49}          & \multicolumn{1}{l}{47.11}            \\ \hline
			\multirow{13}{*}{\rotatebox{90}{\begin{tabular}[c]{@{}l@{}}CSIN with fixed weights \\ (gesture, adversarial, \\and triplet weights)\end{tabular}}} & 0                       & 0.1                     &             60.90                   &              6.14                   &          65.49                     &                 45.48              &                           60.90                   &              6.14                   &          65.49                     &                 45.48                              &                              60.90                   &              6.14                   &          65.49                     &                 45.48                            &                               60.90                   &              6.14                   &          65.49                     &                 45.48                          \\ 
			&                         & 1                      &       60.90     &                            6.14       &                   65.49                                       &                     45.48           &              60.90     &                            6.14       &                   65.49                                       &                     45.48                   &                   60.90     &                            6.14       &                   65.49                                       &                     45.48  &                            60.90     &                            6.14       &                   65.49                                       &                     45.48           \\ 
			&                         & 10                      &                     60.90           & 6.14                                &            65.49                   &       45.48                        &                               60.90           & 6.14                                &            65.49                   &       45.48                               &                               60.90           & 6.14                                &            65.49                   &       45.48                          &                                60.90           & 6.14                                &            65.49                   &       45.48                             \\  \cline{2-19} 
			& \multirow{5}{*}{0.1}    & 0                       &                \underline{61.46}                &              \textbf{5.12}                   &          \textbf{66.14}                     &      \textbf{48.94}                         &                    60.90           &               6.14                  & 65.49 &  45.48                             &          61.34                     &      5.29                           &          65.81                    &       48.07                    &            60.90                    &                6.14                &         65.49                      &     45.48                        \\ 
			&                         & 0.01                    &                      \textbf{61.47}          & \textbf{5.12}                                &               \textbf{66.14}                &         \textbf{48.94}                      &                     60.90          &       6.14                          & 65.49 &  45.48                             &          61.34                     &   5.29                              &        65.81                      &    48.07                       &          61.34                      &              5.29                  &             65.81                  &             48.07                \\ 
			&                         & 0.1                     &           61.09                     &    5.46                             &       65.15                        &   47.69                            &              60.90                 &    6.14                             & 65.49 &  45.48                             &                              61.34 &        5.29                         &    61.47                          &      48.07                     &                  61.34             &          5.29                      &    65.81                           &    48.07                         \\ 
			&                         & 1                      &            61.08                    &    5.44                             &      65.42                         &           47.65                    &            61.19                   &         \underline{5.25}                        & 65.49 &   \underline{48.71}                            &              61.34                 &   5.29                              &         65.81                     &        48.07                   &             61.34                   &                     5.29           &                  65.81             &             48.07                \\ 
			&                         & 10                      &      60.77                          &  6.22                               &        65.26                       &       45.01                        &                 60.90              &         6.14                        &65.49  &   45.48                            &                              61.34 &        5.29                         &     65.81                         &     48.07                      &               61.34                 &         5.29                       &          65.81                     &      48.07                       \\  \cline{2-19} 
			& \multirow{5}{*}{0.01}   & 0                       &          61.00                      &         5.76                        &          65.62                     &         46.73                      &                   60.90            &     6.14                            &  65.49&       45.48                        &              61.34                 &       5.29                          &  65.81                            &     48.07                      &         61.34                       &              5.29                  &                65.81               &           48.07                  \\ 
			&                         & 0.01                   &               61.04                 &  5.69                               &               \underline{65.99}                &        46.75                       &                     60.90          &        6.14                         &  65.49&   45.48                            &                              61.34 &               5.29                  &              65.81                &        48.07                   &                  61.34              &        5.29                        &      65.81                         &  48.07                           \\ 
			&                         & 0.1                     &             61.14                   &    5.43                             &        65.20                       &      47.25                         &         60.90                      &    6.14                             &  65.49&  45.48                             &            61.34                   &           5.29                      &        65.81                      &     48.07                      &                     61.34           &           5.29                     &    65.81                           &        48.07                     \\ 
			&                         & 1                       &              61.02                  &    5.46                             &          65.16                     &        46.99                       &           60.90                    &     6.14                            &65.49  &         45.48                      &               61.34                &         5.29                        &   65.81                           &     48.07                      &                    61.34            &         5.29                       &                              65.81 &           48.07                  \\ 
			&                         & 10                      &          61.14                      &   5.43                              &          65.20                     &       47.25                        &            60.90                   &   6.14                              & 65.49 &         45.48                      &                              61.34 &             5.29                    &         65.81                     &         48.07                  &                 61.34               &             5.29                   &                              65.81 &       48.07                      \\  \bottomrule
		\end{tabular}
	\end{adjustbox}
\end{table*}

\subsection{Performance comparison}

Tables~\ref{tab1} and \ref{tab2} show results against published baselines on UCI EMG and NinaPro DB5, respectively, under leave-one-subject-out cross-validation (LOSOCV). On UCI EMG, the proposed full CSIN reaches 84.48\% mean accuracy, clearing the EMGBench reference of 78.2\% by a meaningful margin. The naïve CSIN (the proposed CSIN with constant Lipschitz) already lands at 80.88\%, which itself surpasses all published comparators. The cost is a slightly wider spread (std 15.74\% vs.\ 15.05\%), driven by a few low-performing subjects where subject-invariant training does not fully close the gap. On NinaPro DB5, the picture is different. Both the naïve CSIN (60.77\%) and full CSIN (61.44\%) substantially outperform EMGBench (41.30\%), and the key gain from the proposed method is not accuracy but variance: std drops from 6.06\% to 1.23\%, and the worst-case subject accuracy rises from 46.00\% to 59.11\%. For a system intended to generalize to unseen users, a nearly five-point improvement in the floor matters more than a fractional gain in the mean.
\par Tables~\ref{tab11} and~\ref{tab22} expand this picture across fixed-weight configurations and four training schedules. On UCI EMG (Table~\ref{tab11}), the baseline (gesture classification only, no adversarial or triplet objectives) sits at 81.75\% (std 14.33\%). Adding triplet loss alone—at any of the three tested weights—barely moves the needle, with means clustering within 0.4\% of baseline regardless of scenario. Adversarial loss alone (with no triplet term) similarly provides negligible lift. The modest gains emerge when both auxiliary objectives are active together. The best fixed-weight result in Table~\ref{tab11} is 82.26\% (std 13.99\%, min 37.76\%) at $\lambda_{\mathrm{adv}}=0.01$ and $\lambda_{\mathrm{trip}}=1$ under Scenario 3—a setup where the subject head is frozen after the warm-up. This pairing sits squarely in the middle of the search grid; notably, raising the triplet weight to 10 rarely helps and occasionally degrades the minimum accuracy, suggesting that aggressive metric pushing conflicts with the classification objective at these embedding scales.
\par Across all four scheduling scenarios the results are strikingly stable. For any fixed pair of $(\lambda_{\mathrm{adv}}, \lambda_{\mathrm{trip}})$, swapping between S0 (fully coupled from epoch one) and S3 (delayed adversarial with later head freezing) typically shifts the mean by under 0.3\%. The practical takeaway is that the benefit does not hinge on a particular initialization trick; the multi-objective formulation is the operative ingredient, not the schedule.

\par On NinaPro DB5 (Table~\ref{tab22}), the baseline (gesture classification only, no adversarial or triplet objectives) is 60.85\% with std 5.60\%. Triplet loss alone leaves performance essentially unchanged. Once adversarial loss is introduced at $\lambda_{\mathrm{adv}}=0.1$, mean accuracy rises to around 61.34--61.47\% across several configurations, and the standard deviation consistently falls to roughly 5.1--5.3\%, compared to 6.14\% without adversarial training. The best individual result, 61.47\% with std 5.12\%, comes at $\lambda_{\mathrm{adv}}=0.1$ and $\lambda_{\mathrm{trip}}=0.01$ under Scenario 0. A lighter adversarial weight of 0.01 $(\lambda_{\mathrm{adv}}=0.01)$ narrows the std gain slightly but still consistently outperforms triplet-only configurations. One structural difference from UCI stands out: on DB5, Scenario 0 (fully coupled training) tends to match or beat the warm-up variants, whereas on UCI the differences were negligible in either direction. With only 10 subjects available for training, the subject head appears to benefit from updating throughout rather than being frozen mid-training—consistent with the intuition that a stronger, continuously updated adversary provides a more persistent invariance signal when number of subjects is less.
\par Taken together, the ablation results point to two clear conclusions. First, neither adversarial training nor triplet loss alone is sufficient; their combination is what drives consistent improvement. Second, the adaptive loss weighting proposed in CSIN removes the need to hand-tune this balance: the Lipschitz-inspired mechanism keeps both auxiliary objectives scaled relative to the classification loss throughout training, which matters especially for datasets where the relative difficulty of invariance and gesture discrimination differs across subjects.

\subsection{Feature extraction progress}
\begin{figure}[!t]
	\centering
	\begin{subfigure}{0.48\linewidth}
		\centering
		\includegraphics[width=\linewidth]{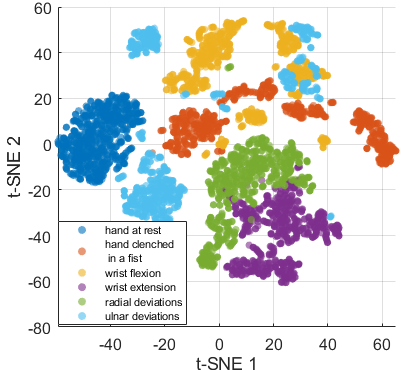}
		\caption{Original data}
		\label{fig:s2}
	\end{subfigure}
	~
	\begin{subfigure}{0.48\linewidth}
		\centering
		\includegraphics[width=\linewidth]{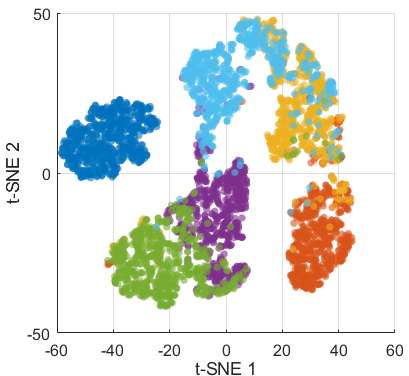}
		\caption{Baseline}
		\label{fig:s2}
	\end{subfigure}
	\hfill
	\begin{subfigure}{0.48\linewidth}
		\centering
		\includegraphics[width=\linewidth]{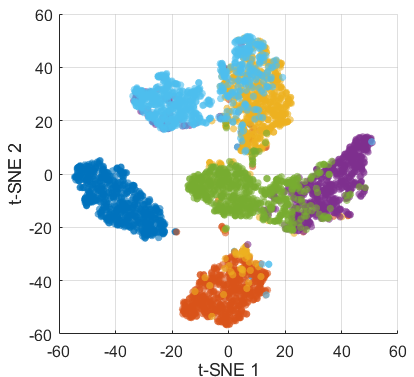}
		\caption{Scenario 1: CSIN with $\lambda_{\mathrm{gest}}=1.0$, $\lambda_{\mathrm{adv}}=0.1$, and $\lambda_{\mathrm{trip}}=0.1$.}
		\label{fig:s2}
	\end{subfigure}
	~
	\begin{subfigure}{0.48\linewidth}
		\centering
		\includegraphics[width=\linewidth]{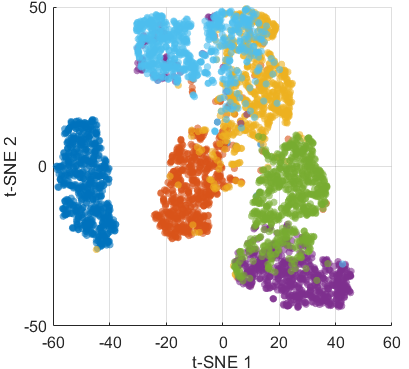}
		\caption{Scenario 1: CSIN with $\lambda_{\mathrm{gest}}=1.0$, $\lambda_{\mathrm{adv}}=0.1$, and $\lambda_{\mathrm{trip}}=10$.}
		\label{fig:s2}
	\end{subfigure}
	\hfill
	\begin{subfigure}{0.48\linewidth}
		\centering
		\includegraphics[width=\linewidth]{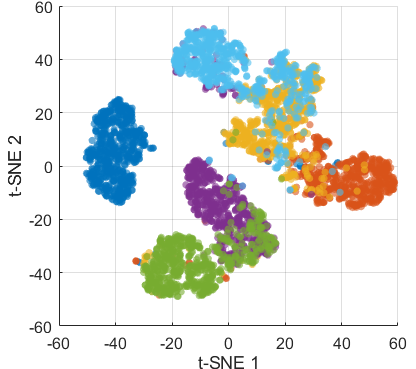}
		\caption{Scenario 2: CSIN with $\lambda_{\mathrm{gest}}=1.0$, $\lambda_{\mathrm{adv}}=0.1$, and $\lambda_{\mathrm{trip}}=0.1$.}
		\label{fig:s2}
	\end{subfigure}
	~
	\begin{subfigure}{0.48\linewidth}
		\centering
		\includegraphics[width=\linewidth]{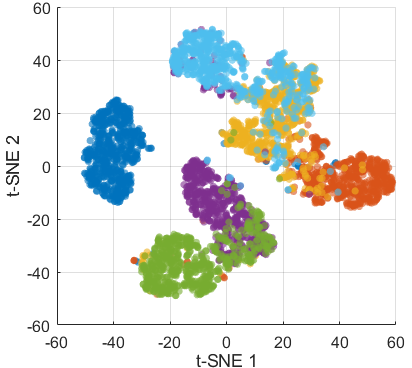}
		\caption{Scenario 2: CSIN with $\lambda_{\mathrm{gest}}=1.0$, $\lambda_{\mathrm{adv}}=0.1$, and $\lambda_{\mathrm{trip}}=10$.}
		\label{fig:s2}
	\end{subfigure}
	\hfill
	\begin{subfigure}{0.48\linewidth}
		\centering
		\includegraphics[width=\linewidth]{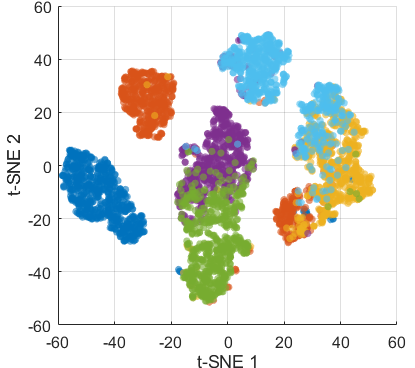}
		\caption{Scenario 3: CSIN with $\lambda_{\mathrm{gest}}=1.0$, $\lambda_{\mathrm{adv}}=0.1$, and $\lambda_{\mathrm{trip}}=0.1$.}
		\label{fig:s2}
	\end{subfigure}
	~
	\begin{subfigure}{0.48\linewidth}
		\centering
		\includegraphics[width=\linewidth]{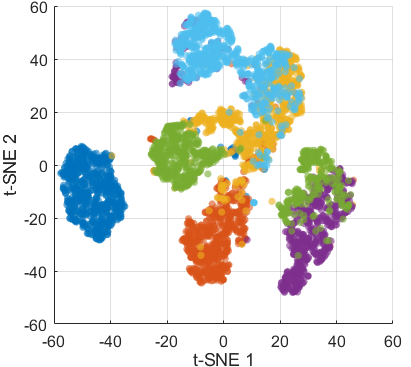}
		\caption{Scenario 3: CSIN with $\lambda_{\mathrm{gest}}=1.0$, $\lambda_{\mathrm{adv}}=0.1$, and $\lambda_{\mathrm{trip}}=10$.}
		\label{fig:s2}
	\end{subfigure}
	\hfill
	\begin{subfigure}{0.48\linewidth}
		\centering
		\includegraphics[width=\linewidth]{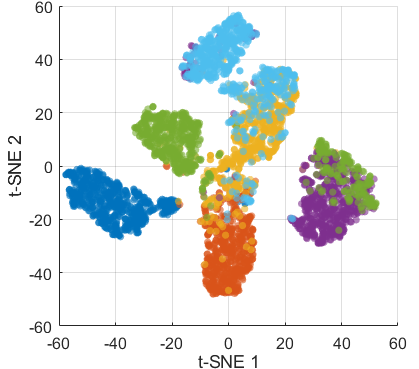}
		\caption{Scenario 4: CSIN with $\lambda_{\mathrm{gest}}=1.0$, $\lambda_{\mathrm{adv}}=0.1$, and $\lambda_{\mathrm{trip}}=0.1$.}
		\label{fig:s2}
	\end{subfigure}
	~
	\begin{subfigure}{0.48\linewidth}
		\centering
		\includegraphics[width=\linewidth]{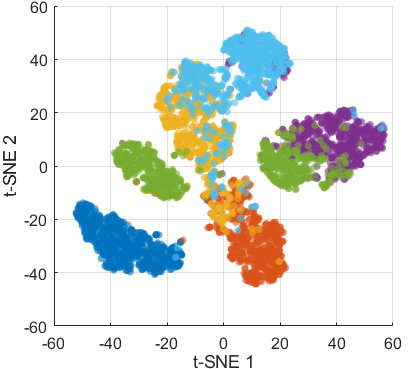}
		\caption{Scenario 4: CSIN with $\lambda_{\mathrm{gest}}=1.0$, $\lambda_{\mathrm{adv}}=0.1$, and $\lambda_{\mathrm{trip}}=10$.}
		\label{fig:s2}
	\end{subfigure}
	\caption{Comparing CSIN under fixed weights and different scenarios.}
	\label{fig:experiment1}
\end{figure}

In this section, we visualized the extracted features under the regular training and the proposed method by t-distributed stochastic neighbor embedding (t-SNE). We trained the two variants with the same architecture on data from 30 subjects and tested them on 6 held-out subjects of the UCI EMG dataset. Fig. \ref{fig:experiment1} shows the visualization result under different loss weights and scenarios. The original feature exhibits a highly scattered, unstructured distribution in the 2D latent space. Except for three gesture classes—wrist extension, radial deviation, and hand at rest—which exhibit relatively well-clustered structures in the original feature space, most other classes are highly scattered and poorly organized, indicating substantial overlap and limited discriminability. Among these, ulnar deviation is the most challenging class, exhibiting the greatest dispersion and the most confusion with other gestures. By comparing regular and adversarial training across all figures, it is evident that the proposed CSIN with fixed weights achieves greater class compactness and stronger inter-class discrimination, particularly for the most challenging gesture, ulnar deviation. Scenario 2 with $\lambda_{\mathrm{trip}}=0.1$ achieves the clearest class separation compared to all other cases and Scenario 4 with $\lambda_{\mathrm{trip}}=10$ provides superior separation among different gestures compared with the other scenarios cases. 

\section{Discussion}\label{sec_Discussion}

\par While our approach deliberately removes subject-specific information to improve cross-subject generalization, there are scenarios where retaining some subject characteristics could actually improve performance. In clinical applications involving patients with neuromuscular disorders or limb differences, subject-specific features may encode crucial information about residual muscle function, compensatory movement patterns, or atypical muscle activation sequences, all of which are essential for accurate gesture recognition. For instance, a stroke patient might consistently activate muscles in an unconventional order to perform a specific gesture, and removing this subject-specific pattern could degrade rather than improve recognition accuracy for that individual. Similarly, in prosthetic control applications where the system is intended for long-term use by a single individual, preserving subject-specific features during an initial calibration phase can lead to more intuitive and responsive control than a completely subject-invariant system. The trade-off becomes particularly relevant when we consider the difference between a universal model that works reasonably well for everyone versus a personalized model that works exceptionally well for one person. In other words, the proposed method could provide a universal baseline in which no after-tuning is applied, and personalization is used.

\par The fundamental challenge we address—removing subject-dependent variations while preserving task-relevant information—extends naturally to other neural recording modalities, though each presents unique considerations. EEG signals exhibit similar cross-subject variability, including differences in skull thickness, brain anatomy, and electrode placement relative to cortical regions, making our adversarial disentanglement approach potentially valuable for motor imagery classification or cognitive state decoding. ECoG and intracortical recordings offer higher signal quality but introduce subject variability due to electrode placement variations, differing tissue responses to implantation, and progressive changes in recording quality over time as the tissue-electrode interface evolves. The similar rules are applied to the other modalities such as KMG \cite{moradi2022clinical}.

\par One finding that deserves more attention than the summary numbers suggest is the behavior of the minimum accuracy across subjects. On UCI EMG, even the best configurations leave certain subjects at 35--38\% accuracy — roughly twice chance level for a 6-class problem, but far enough below the 81\% group average to suggest these individuals are maybe outliers rather than just hard cases at the tail of a normal distribution. This is not a failure of the adversarial mechanism per se; it is a sign that, for some individuals, the four time-domain features used here (MAV, RMS, WL, ZC) may not capture sufficient of the relevant muscle-activation structure to support generalization in either direction. The adversarial loss cannot manufacture a discriminative signal that was never in the features to begin with. This points toward a practical ceiling on what representation-level disentanglement can achieve with a fixed, compact feature set. Learning the features jointly with the encoder, rather than treating them as a fixed preprocessing step, might expose the network to richer signals currently discarded by windowed statistics. The broader point is that invariance learning and feature engineering are not independent problems; in other words, asking the encoder to remove subject-specific information becomes significantly harder when that information is already baked into the input features themselves.

\section{Conclusion}\label{sec_Conclusion}	

Cross-subject generalization in sEMG-based gesture recognition is hard precisely because the signals that identify a person and the signals that identify a gesture are not cleanly separable. This paper addresses that entanglement through a conservative multi-objective framework—CSIN—that jointly optimizes gesture classification, adversarial subject confusion, and triplet-based metric learning on a shared encoder, with a Lipschitz-inspired adaptive mechanism that keeps the auxiliary objectives scaled relative to the primary task throughout training.
\par The experimental results on UCI EMG and NinaPro DB5 support several concrete observations. On UCI EMG, the proposed method reaches 84.48\% mean accuracy under LOSOCV, compared to 78.2\% from the best published comparator. The ablation in Table~\ref{tab11} shows that triplet loss or adversarial loss in isolation provides negligible gain over the classification-only baseline; the improvement requires both objectives active together, with moderate weights. On NinaPro DB5, where only 10 subjects are available and the task involves 10 gestures, the more meaningful gain is in consistency rather than mean accuracy: the standard deviation across subjects drops from 6.06\% to 1.23\%, and the worst-case subject accuracy rises by over 13 points relative to published baselines. Table~\ref{tab22} further shows that the method is robust across training schedules—fully coupled training performs comparably to warm-up variants on UCI, while continuous adversarial updating is slightly preferable on DB5 where number of subjects is limited.
\par Despite these gains, persistent inter-subject variance remains, particularly on UCI EMG where standard deviations stay above 13\% even in the best configurations. Some subjects are simply harder to generalize to, and the current adversarial mechanism does not fully resolve that. Whether this reflects irreducible entanglement between subject identity and gesture-specific activation patterns—or a limitation of the feature set and architecture—is an open question worth pursuing. Extending the framework to richer temporal representations, or incorporating explicit domain randomization during training, could be productive next steps. Applying similar disentanglement strategies to EEG or intracortical recordings, where analogous cross-subject variability challenges exist, is another natural direction for future work.
\section*{Code availability}
All results and codes will be available at \url{www.github.com/Hamed-Rafiei/CSIN}.
\section*{Data availability}
All datasets used in this paper are publicly available.

\bibliographystyle{ieeetr}

\bibliography{mybibs}

\begin{thebibliography}{10}

\bibitem{oskoei2007myoelectric}
M.~A. Oskoei and H.~Hu, ``Myoelectric control systems—a survey,'' {\em
  Biomedical Signal Processing and Control}, vol.~2, no.~4, pp.~275--294, 2007.

\bibitem{ge2023gesture}
Z.~Ge, Z.~Wu, X.~Han, and P.~Zhao, ``Gesture recognition and master--slave
  control of a manipulator based on semg and convolutional neural
  network--gated recurrent unit,'' {\em Journal of Engineering and Science in
  Medical Diagnostics and Therapy}, vol.~6, no.~2, p.~021004, 2023.

\bibitem{jaramillo2020real}
A.~Jaramillo-Y{\'a}nez, M.~E. Benalc{\'a}zar, and E.~Mena-Maldonado,
  ``Real-time hand gesture recognition using surface electromyography and
  machine learning: {A} systematic literature review,'' {\em Sensors}, vol.~20,
  no.~9, p.~2467, 2020.

\bibitem{zhang2024online}
Z.~Zhang, S.~Liu, Y.~Wang, W.~Song, and Y.~Zhang, ``Online cross session
  electromyographic hand gesture recognition using deep learning and transfer
  learning,'' {\em Engineering Applications of Artificial Intelligence},
  vol.~127, p.~107251, 2024.

\bibitem{rafiei2025understandable}
H.~Rafiei and M.-R. Akbarzadeh-T, ``Understandable time frame-based biosignal
  processing,'' {\em Biomedical Signal Processing and Control}, vol.~103,
  p.~107429, 2025.

\bibitem{liu2015towards}
J.~Liu, X.~Sheng, D.~Zhang, N.~Jiang, and X.~Zhu, ``Towards zero retraining for
  myoelectric control based on common model component analysis,'' {\em IEEE
  Transactions on Neural Systems and Rehabilitation Engineering}, vol.~24,
  no.~4, pp.~444--454, 2015.

\bibitem{du2017surface}
Y.~Du, W.~Jin, W.~Wei, Y.~Hu, and W.~Geng, ``Surface emg-based inter-session
  gesture recognition enhanced by deep domain adaptation,'' {\em Sensors},
  vol.~17, no.~3, p.~458, 2017.

\bibitem{pradhan2021performance}
A.~Pradhan, J.~He, and N.~Jiang, ``Performance optimization of surface
  electromyography based biometric sensing system for both verification and
  identification,'' {\em IEEE Sensors Journal}, vol.~21, no.~19,
  pp.~21718--21729, 2021.

\bibitem{jiang2022optimization}
X.~Jiang, X.~Liu, J.~Fan, X.~Ye, C.~Dai, E.~A. Clancy, D.~Farina, and W.~Chen,
  ``Optimization of hd-semg-based cross-day hand gesture classification by
  optimal feature extraction and data augmentation,'' {\em IEEE Transactions on
  Human-Machine Systems}, vol.~52, no.~6, pp.~1281--1291, 2022.

\bibitem{yang2024emgbench}
J.~Yang, M.~Soh, V.~Lieu, D.~J. Weber, and Z.~Erickson, ``Emgbench:
  {Benchmarking} out-of-distribution generalization and adaptation for
  electromyography,'' {\em Advances in Neural Information Processing Systems},
  vol.~37, pp.~50313--50342, 2024.

\bibitem{fan2024surface}
J.~Fan, X.~Jiang, X.~Liu, L.~Meng, F.~Jia, and C.~Dai, ``Surface emg feature
  disentanglement for robust pattern recognition,'' {\em Expert Systems with
  Applications}, vol.~237, p.~121224, 2024.

\bibitem{phinyomark2013emg}
A.~Phinyomark, F.~Quaine, S.~Charbonnier, C.~Serviere, F.~Tarpin-Bernard, and
  Y.~Laurillau, ``Emg feature evaluation for improving myoelectric pattern
  recognition robustness,'' {\em Expert Systems with Applications}, vol.~40,
  no.~12, pp.~4832--4840, 2013.

\bibitem{fan2023improving}
J.~Fan, M.~Jiang, C.~Lin, G.~Li, J.~Fiaidhi, C.~Ma, and W.~Wu, ``Improving
  semg-based motion intention recognition for upper-limb amputees using
  transfer learning,'' {\em Neural Computing and Applications}, vol.~35,
  no.~22, pp.~16101--16111, 2023.

\bibitem{wang2023iterative}
K.~Wang, Y.~Chen, Y.~Zhang, X.~Yang, and C.~Hu, ``Iterative self-training based
  domain adaptation for cross-user semg gesture recognition,'' {\em IEEE
  Transactions on Neural Systems and Rehabilitation Engineering}, vol.~31,
  pp.~2974--2987, 2023.

\bibitem{su2025multi}
K.~Su, K.~Liu, B.~Wan, H.~Qiao, J.~Huang, M.~Feng, and J.~Liu, ``Multi-source
  adversarial feature disentanglement method for cross-subject gesture
  recognition using semg signals,'' {\em IEEE Transactions on Instrumentation
  and Measurement}, vol.~74, pp.~1--12, 2025.

\bibitem{neacsu2024lipschitz}
A.~A. Neac{\c{s}}u, J.-C. Pesquet, and C.~Burileanu, ``Emg-based automatic
  gesture recognition using lipschitz-regularized neural networks,'' {\em ACM
  Transactions on Intelligent Systems and Technology}, vol.~15, no.~2,
  p.~Article 26, 2024.

\bibitem{tsuzuku2018lipschitz}
Y.~Tsuzuku, I.~Sato, and M.~Sugiyama, ``Lipschitz-margin training: Scalable
  certification of perturbation invariance,'' in {\em NeurIPS}, 2018.

\bibitem{gouk2021regularisation}
H.~e.~a. Gouk, ``Regularisation of neural networks by enforcing lipschitz
  continuity,'' in {\em Machine Learning}, 2021.

\bibitem{gao2022multifeatured}
Z.~Gao, Y.~Wang, X.~Sun, P.~Chen, and C.~Ma, ``A multifeatured time--frequency
  neural network system for classifying semg,'' {\em IEEE Transactions on
  Circuits and Systems II: Express Briefs}, vol.~69, no.~11, pp.~4588--4592,
  2022.

\bibitem{shen2023ica}
S.~Shen, X.~Wang, M.~Wu, K.~Gu, X.~Chen, and X.~Geng, ``Ica-cnn: {Gesture}
  recognition using cnn with improved channel attention mechanism and
  multimodal signals,'' {\em IEEE Sensors Journal}, vol.~23, no.~4,
  pp.~4052--4059, 2023.

\bibitem{zhang2020learning}
Y.~Zhang, Y.~Chen, H.~Yu, X.~Yang, and W.~Lu, ``Learning effective
  spatial--temporal features for semg armband-based gesture recognition,'' {\em
  IEEE Internet of Things Journal}, vol.~7, no.~8, pp.~6979--6992, 2020.

\bibitem{chen2020hand}
X.~Chen, Y.~Li, R.~Hu, X.~Zhang, and X.~Chen, ``Hand gesture recognition based
  on surface electromyography using convolutional neural network with transfer
  learning method,'' {\em IEEE Journal of Biomedical and Health Informatics},
  vol.~25, no.~4, pp.~1292--1304, 2020.

\bibitem{wang2025residual}
H.~Wang, D.~Jiang, J.~Yun, Y.~Zhao, L.~Huang, Y.~Liu, M.~Jia, and B.~Chen,
  ``Residual attention-based hybrid neural network for semg gesture
  recognition,'' {\em Concurrency and Computation: Practice and Experience},
  vol.~37, no.~25-26, p.~e70368, 2025.

\bibitem{cote2020interpreting}
U.~C{\^o}t{\'e}-Allard, E.~Campbell, A.~Phinyomark, F.~Laviolette, B.~Gosselin,
  and E.~Scheme, ``Interpreting deep learning features for myoelectric control:
  {A} comparison with handcrafted features,'' {\em Frontiers in Bioengineering
  and Biotechnology}, vol.~8, p.~158, 2020.

\bibitem{zhang2025extended}
Z.~Zhang, Y.~Ming, Q.~Shen, Y.~Wang, and Y.~Zhang, ``An extended variational
  autoencoder for cross-subject electromyograph gesture recognition,'' {\em
  Biomedical Signal Processing and Control}, vol.~99, p.~106828, 2025.

\bibitem{pizzolato2017comparison}
S.~Pizzolato, L.~Tagliapietra, M.~Cognolato, M.~Reggiani, H.~M{\"u}ller, and
  M.~Atzori, ``Comparison of six electromyography acquisition setups on hand
  movement classification tasks,'' {\em PloS one}, vol.~12, no.~10,
  p.~e0186132, 2017.

\bibitem{smith2010determining}
L.~H. Smith, L.~J. Hargrove, B.~A. Lock, and T.~A. Kuiken, ``Determining the
  optimal window length for pattern recognition-based myoelectric control:
  {Balancing} the competing effects of classification error and controller
  delay,'' {\em IEEE Transactions on Neural Systems and Rehabilitation
  Engineering}, vol.~19, no.~2, pp.~186--192, 2010.

\bibitem{lee2024decoding}
H.~Lee, M.~Jiang, J.~Yang, Z.~Yang, and Q.~Zhao, ``Decoding gestures in
  electromyography: {Spatiotemporal} graph neural networks for generalizable
  and interpretable classification,'' {\em IEEE Transactions on Neural Systems
  and Rehabilitation Engineering}, vol.~33, pp.~404--419, 2024.

\bibitem{lee2025fedemg}
H.~Lee, M.~Jiang, and Q.~Zhao, ``Fedemg: Achieving generalization,
  personalization, and resource efficiency in emg-based upper-limb
  rehabilitation through federated prototype learning,'' {\em IEEE Transactions
  on Biomedical Engineering}, vol.~73, pp.~803--817, 2025.

\bibitem{hoshino2024comparison}
T.~Hoshino, S.~Kanoga, M.~Tsubaki, and A.~Aoyama, ``Comparison of fine-tuned
  single-source and multi-source approaches to surface electromyogram pattern
  recognition,'' {\em Biomedical Signal Processing and Control}, vol.~94,
  p.~106261, 2024.

\bibitem{moradi2022clinical}
A.~Moradi, H.~Rafiei, M.~Daliri, M.-R. Akbarzadeh-T, A.~Akbarzadeh, A.-M.
  Naddaf-Sh, and S.~Naddaf-Sh, ``Clinical implementation of a bionic hand
  controlled with kineticomyographic signals,'' {\em Scientific Reports},
  vol.~12, no.~1, p.~14805, 2022.

\end{thebibliography}

\end{document}